# An Open-Source, Event-Driven Pipeline for Cryptocurrency Market Data: Ingestion, Forecasting, and On-Chain Fraud Detection

Basil Sajid Shaikh, shaikh.basil786@gmail.com ,Seattle, WA, USA

Melrick Mascarenhas, melrickjm@gmail.com, San Diego, CA, USA

Dr. Nuzhat Faiz Shaikh, nfshaikh@mescoepune.org, Pune, India

## Abstract

Cryptocurrency markets generate high-frequency, multi-source data that is expensive to work with unless a team already has commercial-grade streaming and warehousing infrastructure in place. This paper describes a fully open-source pipeline that reproduces the behavior of a cloud-native, event-driven system — file arrival triggering a message, a message triggering compute — entirely on commodity hardware, using Apache Kafka and a filesystem-watching poller in place of managed cloud triggers. The pipeline partitions historical Gemini exchange data into hourly and minutely files, ingests them asynchronously through two independently grouped Kafka consumers (one for audit logging, one for Spark-triggered ETL), and lands cleaned output in a PostgreSQL warehouse with historical and aggregated schemas plus asset-specific data marts. We use the resulting Bitcoin data mart to compare a seasonal ARIMA model against a single-layer LSTM network for price forecasting, and separately apply Random Forest and Gradient Boosting classifiers, with additional engineered features, to the public Ethereum fraud detection benchmark introduced by Farrugia et al. We report the architecture, the modeling methodology, and the resulting metrics, and we are explicit about the limitations of comparing forecasts issued at different horizons and of evaluating fraud detection on a static, already-labeled dataset.

## 1. Introduction

Thousands of cryptocurrencies now trade around the clock across dozens of exchanges, and each one produces a continuous stream of order and price data. Working with that data at any serious scale has traditionally meant either paying for a managed streaming and warehousing stack, or building something bespoke and hoping it holds together. Neither option is very accessible to a small research group, a fintech startup validating an idea, or a student project that still needs to look and behave like production infrastructure. This paper is about building the second option properly: a pipeline that behaves like a cloud-native, event-driven system without actually renting one.

The design goal was specific: reproduce the trigger pattern that a managed cloud setup gives you for free — a file lands in storage, that arrival fires an event, the event kicks off compute — using only open-source components running on a single machine. We built this around Apache Kafka and a lightweight filesystem watcher standing in for cloud storage triggers, Apache Spark for the transformation layer, and PostgreSQL for storage. To check that the resulting warehouse is actually useful and not just a working demo, we ran two independent modeling exercises against it: a comparison of classical and deep-learning forecasting methods on Bitcoin prices, and a wallet-level fraud classifier built on a public Ethereum benchmark. The second exercise uses external, previously labeled data rather than data produced by the pipeline itself, and we are careful throughout to keep that distinction visible rather than implying a tighter integration than exists.

This paper makes four contributions:

- An event-driven ingestion design, built entirely from open-source components, that reproduces the storage-trigger-compute pattern of a managed cloud stack without any cloud spend, using two independent Kafka consumer groups reading the same topic to separate audit logging from compute-triggering.

- A two-schema warehouse and per-asset data mart layer on PostgreSQL that supports both dashboarding and downstream modeling from the same cleaned data, avoiding a rebuild of the transformation logic for each new consumer.
- A comparison of a seasonal ARIMA model and an LSTM network for Bitcoin price forecasting, evaluated on data drawn from the pipeline's own output, along with an explicit accounting of why the two evaluations are not on equal footing.
- A replication and light extension of the Farrugia et al. Ethereum illicit-account benchmark, adding eight engineered ratio and diversity features and comparing Random Forest against Gradient Boosting.

# 2. Related Work

Event-driven ingestion of this kind traces back to Kafka's original design as a high-throughput, partition-based log for decoupling producers from consumers at LinkedIn [1]. The pattern we use — a lightweight process publishing metadata about new files, with downstream consumers reading independently and at their own pace — is a direct, locally executed analogue of that design, substituting a filesystem watcher for the managed storage-trigger services a cloud deployment would normally use. The Spark side of the pipeline builds on the resilient distributed dataset abstraction introduced by Zaharia et al., which is what makes it practical to merge, clean, and aggregate many small files in parallel rather than processing them one at a time [2].

On the forecasting side, ARIMA and LSTM are both well-studied for cryptocurrency price prediction, and the general finding across this literature is consistent: LSTM and other non-linear sequence models tend to track short-horizon price movement more closely than ARIMA-family models, though ARIMA remains a fast, interpretable baseline. McNally et al. compared Bayesian-optimized RNN and LSTM models against ARIMA for Bitcoin and found the deep learning approaches outperformed the classical baseline on direction and error [3]. Siami-Namini et al. reached a similar conclusion across a broader set of financial time series, reporting substantially lower forecast error for LSTM relative to ARIMA [4]. Our own comparison in Section 5 lines up qualitatively with these results, though as we note there, our two models were not evaluated under identical conditions, so the comparison should be read as indicative rather than as a controlled benchmark.

On the fraud detection side, the dataset we use was introduced by Farrugia, Ellul, and Azzopardi, who collected account-level transactional and behavioral features for Ethereum addresses flagged by the community as illicit, alongside a matched set of regular accounts, and trained an XGBoost classifier reaching 96.3% accuracy and a 99.4% AUC [5]. That dataset and feature set have since been reused by a number of follow-up studies applying different model families and additional feature engineering. Section 6 follows the same general approach — the same public dataset, a similar feature-engineering philosophy — but tests Random Forest and Gradient Boosting rather than XGBoost, with eight additional engineered features. We report where our results land relative to the original benchmark, while being clear that differences in train/test splitting and preprocessing mean the comparison is suggestive rather than strictly controlled.

Most existing write-ups of this kind cover either the systems side (a streaming architecture demonstration) or the modeling side (a forecasting or fraud-detection study) in isolation. This paper documents both halves of one shared pipeline, and is explicit that the Ethereum dataset is an external benchmark rather than a product of the ingestion system, so as not to overstate how integrated the two halves are.

# 3. System Architecture

This section describes the pipeline end to end, from raw historical files to the schemas that downstream consumers query. Each subsystem is described along with the specific design decisions behind it, since those decisions — not the individual tools — are what make the architecture reproducible on a single machine and, as Section 4 argues, extensible beyond it.

## 3.1 Data acquisition and partitioning

Historical OHLCV (open, high, low, close, volume) trading data for the Gemini exchange was obtained from CryptoDataDownload at hourly resolution, covering multiple assets (Bitcoin, Ethereum, and others) as separate CSV files with a common schema: a Unix timestamp, a human-readable date, the trading pair symbol, open/high/low/close prices, and base- and quote-currency volume. Because the goal was to simulate incremental, daily arrival of data rather than a single bulk load, a script (toDailyData.py) reads each historical file and splits it into one file per trading day for a target date range. During this split, rows with a missing timestamp, a non-positive price, or a volume field that fails to parse as numeric are discarded, and every retained timestamp is normalized to ISO 8601 so that later stages never have to reason about mixed date formats. The output is written into two directories, S3_Hourly/ and S3_Minutely/, organized by asset and date (for example, S3_Hourly/BTCUSD/2017-05-15.csv), which act as a local stand-in for a cloud object store and give the rest of the pipeline a realistic, incrementally growing input to react to.

## 3.2 Event-driven ingestion

A Watchdog-based Python poller (watchAndPublish.py) monitors both directories using the operating system's native filesystem-event API (inotify on Linux) rather than a fixed-interval scan, so a new file is detected within milliseconds of being written rather than on the next poll cycle. On each creation or modification event it publishes a small JSON message to a Kafka topic (crypto-file-events):

```
{ "mode": "created", "type": "hourly", "path": "S3_Hourly/BTCUSD/2017-05-15.csv",
"detected_at": "2017-05-16T00:00:04Z" }
```

The message carries only metadata, not the file's contents — consumers read the file themselves once notified, which keeps individual Kafka messages small and keeps the broker from becoming a bottleneck as file sizes grow. This is the piece that stands in for an object-storage trigger invoking a serverless function: the same decoupling of detection from processing, running locally at no infrastructure cost. Because the topic is partitioned by asset symbol, events for different assets can be consumed and processed in parallel rather than being serialized through a single queue.

## 3.3 Two independent Kafka consumers

Two consumer groups read the same topic independently, which is the central design choice of the ingestion layer. The first (log_data_driver.py) is a lightweight logging consumer that appends every event it sees, with a timestamp and its own processing latency, to an audit log; it exists purely for operational visibility, has no downstream dependents, and introduces no processing delay for anything else. The second (consumer.py) buffers incoming events by type in memory and triggers a Spark job once a threshold is reached — 28 files for a full day of hourly data, or 5 files for a smaller minutely test batch. Because the groups are independent, Kafka tracks a separate committed offset for each: if the Spark-triggering consumer crashes mid-batch, it resumes from its last committed offset on restart without reprocessing already-handled files and without the logging consumer ever noticing the interruption. Adding a third consumer — for alerting on abnormal file sizes, for instance, or for feeding a second warehouse — requires no change to either of the existing two, since each consumer group sees the full, independent event stream.

## 3.4 Spark-based ETL and the warehouse

Once triggered, the Spark job reads the batch of files into a single DataFrame, drops rows with null or negative values in the timestamp, price, or volume fields, casts every field to its proper type, and aggregates the batch into daily summaries (mean, median, minimum, and maximum price, and total volume). The job runs in local mode for this pipeline, but the transformation logic is expressed entirely in Spark's DataFrame API rather than in row-by-row Python, so the same code runs unmodified against a multi-node cluster if the input volume outgrows a single machine. The output is written to a PostgreSQL instance on AWS RDS in two schemas: a historical schema holding the cleaned, full-resolution records (timestamp, symbol, open, high, low, close, volume) with a composite index on symbol and timestamp for fast range queries, and an aggregated schema holding the daily rollups used for dashboarding and modeling. Writes use an upsert (insert-or-update) on the timestamp-symbol key, so a file that is reprocessed — after a crash recovery, for instance — overwrites its own prior output instead of creating duplicate rows.

## 3.5 Data marts and dashboards

A SQL-driven mart layer extracts an asset-specific slice of the historical schema — currently Bitcoin — into its own schema, so that downstream consumers query a small, relevant table rather than the full warehouse. The mart is refreshed by a scheduled query rather than a full recompute, so its cost stays proportional to new data rather than to the warehouse's total size. Two Tableau dashboards sit on top of this layer: one for single-asset exploration (price, volume, moving averages) and one for side-by-side comparison across multiple cryptocurrencies. Figure 1 shows the complete flow from raw historical files through to these downstream consumers, including the Ethereum fraud-detection dataset, which is drawn independently.

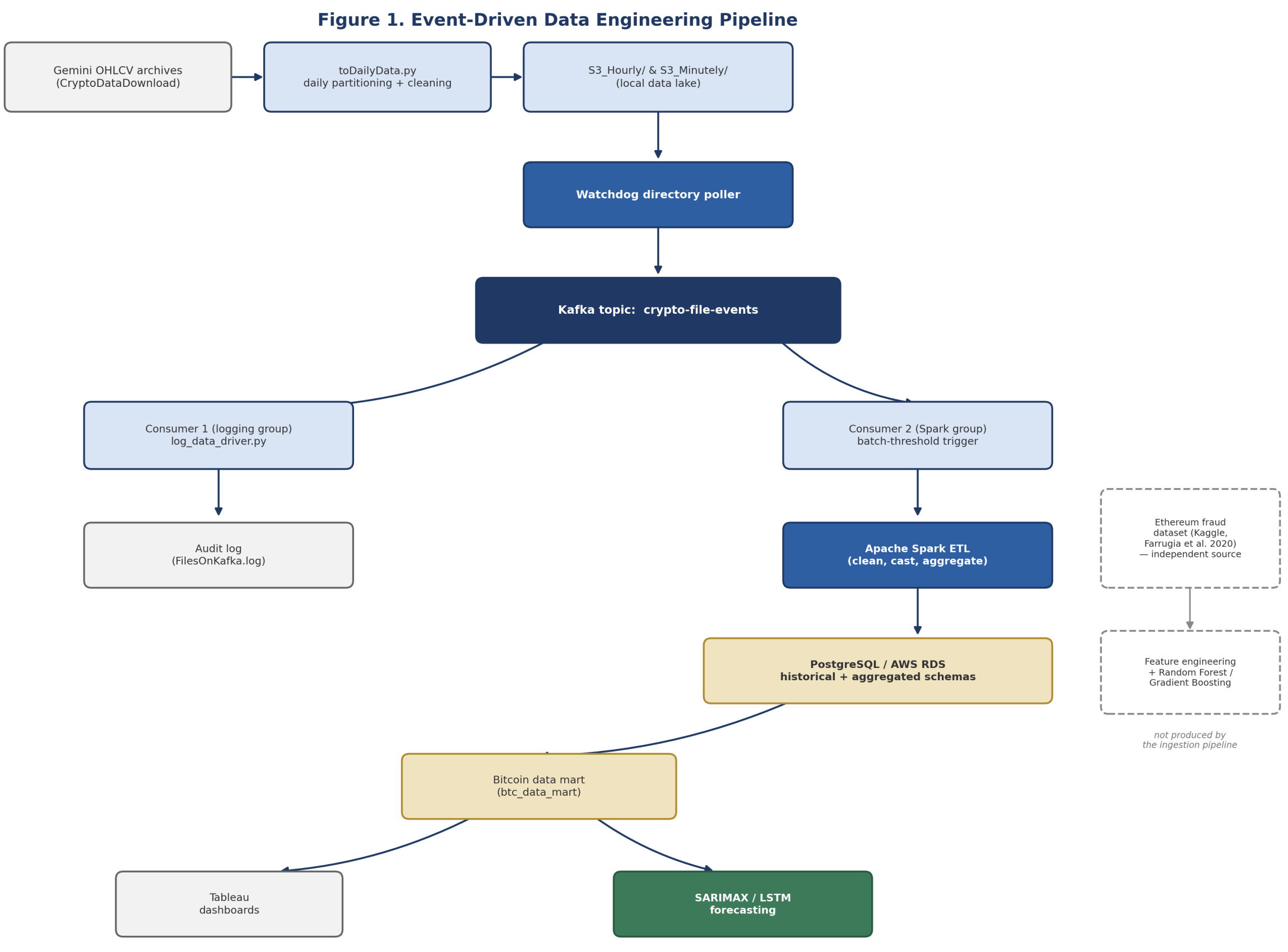

*Figure 1. End-to-end architecture. The Ethereum fraud dataset (dashed boxes) is an external, publicly available benchmark and is not produced by this ingestion pipeline.*

# 4. Architectural Advantages: Scalability and Extensibility

The individual components of this pipeline — Kafka, Spark, PostgreSQL, Tableau — are all standard choices; what we think is worth stating explicitly is why assembling them this way, with this specific separation of concerns, pays off as the system grows. Four properties fall out of the design directly.

## 4.1 Independent scaling of each stage

Because ingestion, transformation, and storage are three separate systems connected by a message queue rather than one monolithic script, each can be scaled on its own terms. If file arrival rate increases, the Kafka topic can be repartitioned and the ingestion side scaled out with more consumer instances in the same group, with Kafka distributing partitions across them automatically. If the transformation workload grows, the Spark job can move from local mode to a multi-node cluster without any change to the ETL logic itself, since that logic is already expressed in Spark's distributed DataFrame API rather than in single-threaded code. If read load on the warehouse grows — more dashboards, more modeling jobs — read replicas can be added to the PostgreSQL/RDS layer without touching ingestion or transformation at all. None of these scaling moves requires redeploying or even restarting the other stages, because the stages only communicate through the topic and the warehouse, never directly.

## 4.2 Fault isolation

The dual-consumer-group pattern described in Section 3.3 means a failure in one path cannot propagate to the other: the audit-logging consumer can fall behind or crash without ever affecting the Spark-triggering path that actually produces warehouse data, and vice versa. More generally, because every stage commits its own progress (Kafka offsets for consumers, upsert keys for warehouse writes), a crash at any point in the pipeline is recoverable by restarting that stage alone — nothing needs to be replayed from the beginning, and nothing downstream needs to be told anything happened.

### 4.3 Horizontal extensibility to new assets and data sources

Onboarding a new tradable asset does not require new code in the ingestion or transformation layers. Because the watcher, the Kafka topic, and the Spark job all key on the asset symbol carried in the file path and the event payload, dropping a new asset's daily files into the watched directories is sufficient for it to flow through the existing pipeline; only a new asset-specific data mart (Section 3.5) needs to be added to expose it to a dedicated dashboard or model. The same reasoning applies to adding an entirely new upstream data source — a different exchange, or a different data vendor — since the only requirement on that source is that it can be normalized into the same partitioned-file convention the watcher already monitors. This is also why the architecture cleanly accommodates a dataset like the Ethereum fraud benchmark in Section 6 sitting alongside it: a new downstream consumer of the warehouse (or, as in that case, an entirely separate benchmark) can be added without modifying the ingestion or transformation code at all.

### 4.4 Extensibility of downstream consumption

Splitting the warehouse into a historical schema and an aggregated schema, and further into per-asset marts, means new downstream consumers do not have to re-derive cleaned data from raw files themselves. The forecasting experiments in Section 5 and the dashboard layer in Section 3.5 both read from the same Bitcoin mart without any coordination between them or any duplicated cleaning logic; a future fraud-detection model built directly on this pipeline's own transaction data, rather than on the external Ethereum benchmark used in Section 6, would plug into the same warehouse the same way. In practice this means the marginal cost of adding a new analytical use case is a new read-only query against an existing schema, not a new ingestion path.

Table 1 summarizes these properties against the two more conventional alternatives — a single bulk-load ETL script, and a fully managed cloud pipeline (e.g., S3 event notifications, Lambda, and a managed warehouse) — to make the trade-offs explicit rather than implicit.

*Table 1. Qualitative comparison of ingestion strategies for this workload.*

| Property | Single bulk-load script | Fully managed cloud (S3 + Lambda) | This pipeline (local, event-driven) |
|---|---|---|---|
| Infrastructure cost | Low (single script) | Pay-per-use, scales with volume | Zero cloud spend; commodity hardware |
| Reaction latency to new data | None — batch only | Near-real-time (managed trigger) | Near-real-time (OS filesystem events) |
| Fault isolation between stages | None — one process | Strong (managed services) | Strong (independent consumer groups) |
| Horizontal scaling path | Rewrite required | Automatic, managed | Manual but code-compatible (Kafka partitions, Spark cluster) |
| Adding a new asset or source | Modify the script | Add a bucket + trigger rule | Drop files into a watched directory |
| Operational transparency | Print statements / ad hoc logs | Cloud provider console | Independent, queryable audit log |

## 5. Forecasting Bitcoin Prices

We evaluated two forecasting approaches against Bitcoin data drawn from the pipeline's warehouse: a seasonal ARIMA model as a classical, interpretable baseline, and an LSTM network to capture non-linear dynamics. The two models were developed against different resamplings of the underlying data — monthly for ARIMA over a roughly five-year span, daily for LSTM over a more recent window — a discrepancy we return to in Section 5.3.

## 5.1 SARIMAX baseline

Minute-level Bitcoin data from the data mart was resampled to monthly averages, using a weighted average of the high, low, and close prices for each interval. An Augmented Dickey-Fuller (ADF) test on the raw series returned a p-value of 0.998, confirming non-stationarity. A Box-Cox transform to stabilize variance, followed by seasonal and first-order differencing, brought the ADF p-value down to 0.045, at which point the series was treated as stationary. Figure 2 shows the seasonal decomposition of the series before and after this transformation.

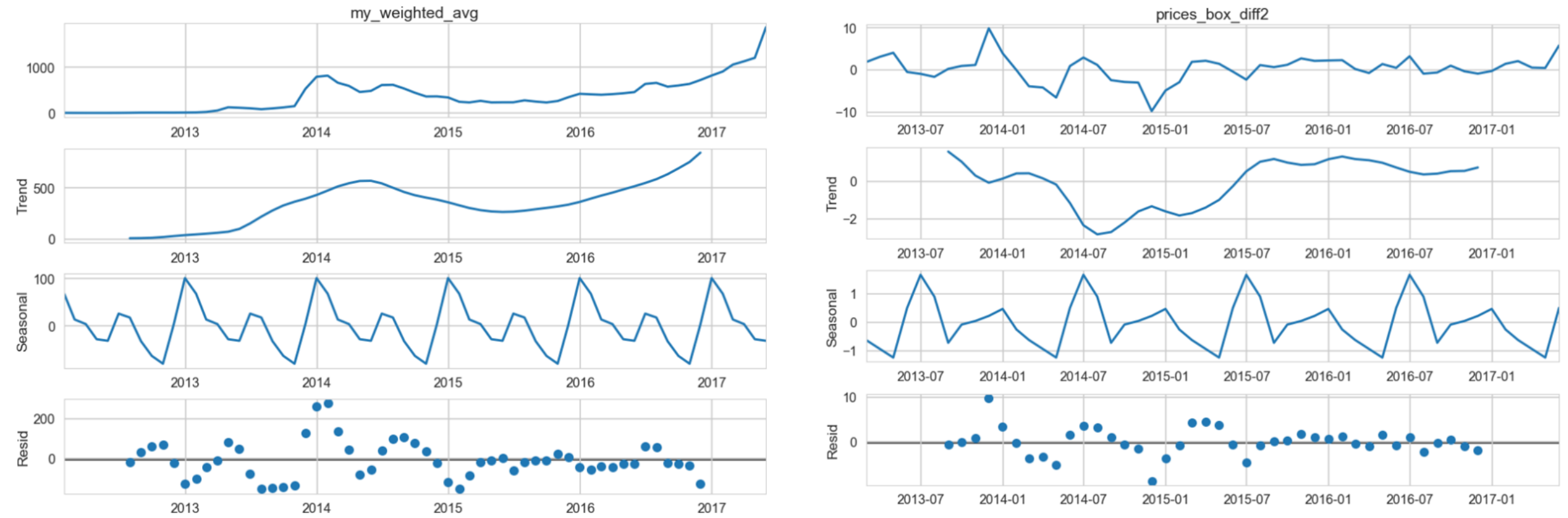

*Figure 2. Seasonal decomposition of the weighted-average Bitcoin price before (left) and after (right) Box-Cox transformation and differencing. The right panel's residual component is visibly closer to noise.*

A grid search over seasonal and non-seasonal (p, d, q) combinations, scored by Akaike Information Criterion (AIC), selected SARIMAX(1,1,0)×(0,1,1,12) with an AIC of 231.86. Table 2 lists the fitted parameters. The residual series passed a second ADF test with a p-value under 0.001, and its autocorrelation function showed no significant remaining structure, indicating the model had captured the available temporal pattern.

*Table 2. SARIMAX(1,1,0)×(0,1,1,12) parameter estimates (n = 65 monthly observations, AIC = 231.86, BIC = 237.72).*

| Parameter | Coefficient | Std. err. | z | P>\|z\| |
|---|---|---|---|---|
| ar.L1 | 0.4830 | 0.189 | 2.557 | 0.011 |
| ma.S.L12 | −0.9937 | 9.889 | −0.100 | 0.920 |
| sigma2 | 3.0677 | 30.123 | 0.102 | 0.919 |

Forecasts were generated six months beyond the training window and transformed back to the original price scale via the inverse Box-Cox transform. Figure 3 compares the forecast against the actual series; the model tracks the broader seasonal cycle and upward trend reasonably well, though, as expected of a linear model, it does not anticipate the sharp late-sample acceleration in price.

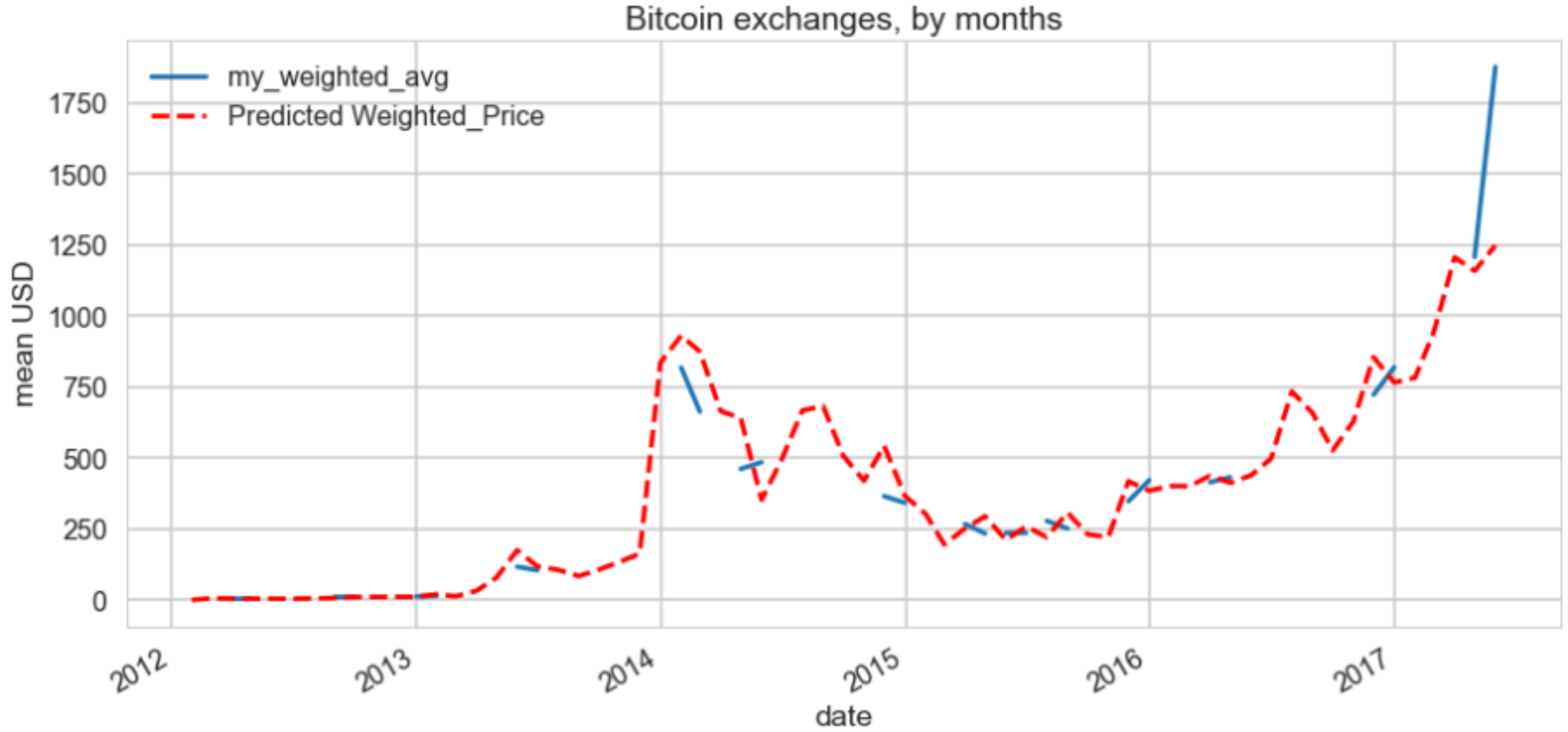

*Figure 3. SARIMAX six-month-ahead forecast versus the actual weighted-average monthly price.*

## 5.2 LSTM network

For the LSTM study, the same weighted-average price definition, (high + low + close) / 3, was resampled to a daily frequency and scaled to [0, 1] with a min-max scaler. The task was framed as one-step-ahead prediction: given day t's price, predict day t + 1. The network itself is deliberately small — a single LSTM layer with four hidden units and a sigmoid activation, followed by a one-unit dense output layer — trained with the Adam optimizer against mean squared error for 100 epochs with a batch size of 5. The final 30 days of the series were held out for testing. Figure 4 shows predicted versus actual prices on this held-out window; the model tracks short-term movement closely despite its small size, though it lags slightly during the sharpest swings.

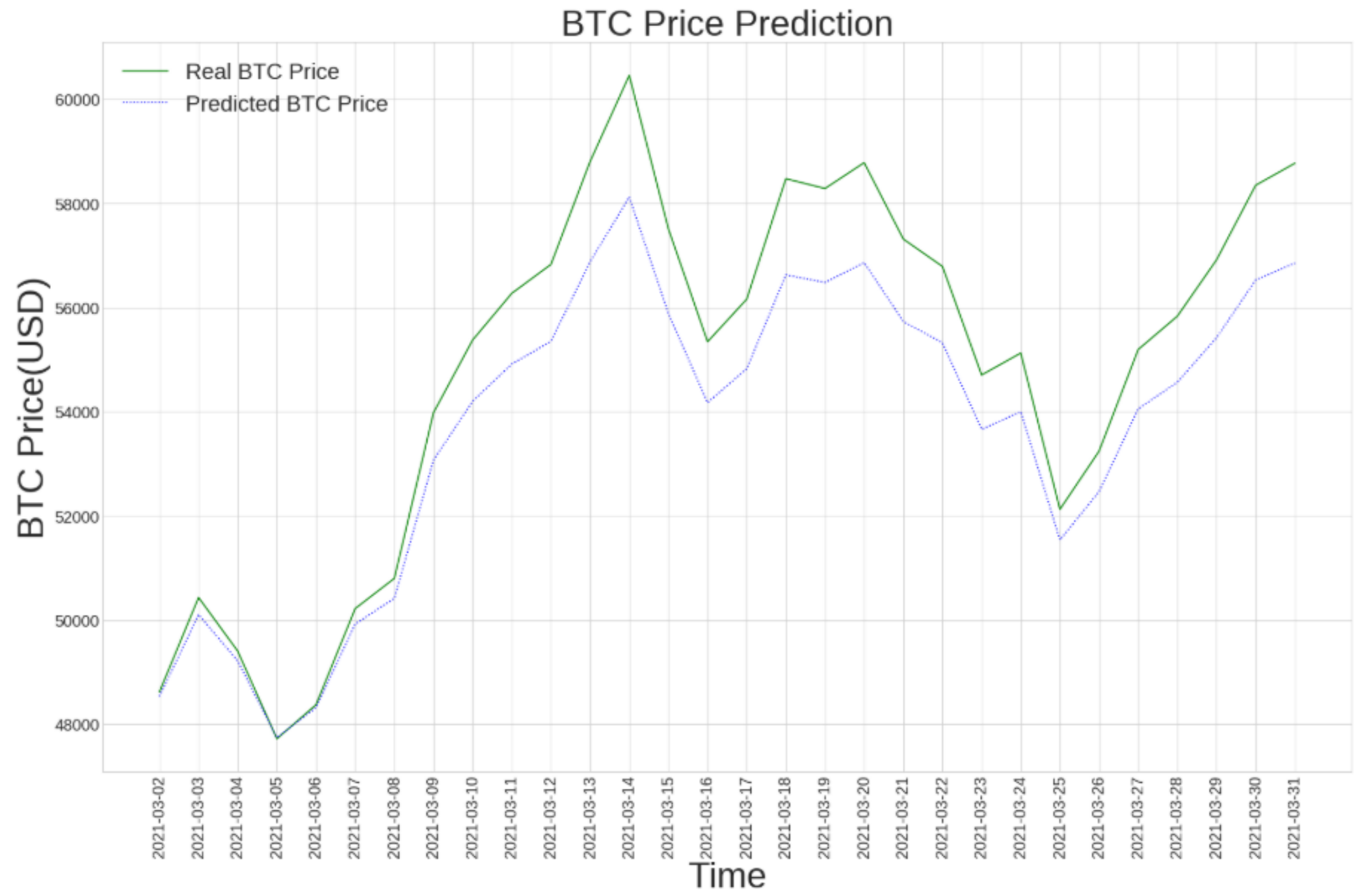

*Figure 4. LSTM one-step-ahead predictions versus actual Bitcoin prices over the 30-day test window.*

## 5.3 Comparison and its limits

Table 3 reports RMSE and MAE for both models. LSTM shows substantially lower error than SARIMAX on its own test window, consistent with the broader literature on ARIMA versus LSTM for cryptocurrency forecasting [3], [4]. That said, the two rows in Table 3 are not a controlled comparison: they were fit on different resamplings (monthly versus daily), over different historical spans, and evaluated at different forecast horizons (six months versus 30 days). We report both because each was the setup used in its own study, but the correct reading of Table 3 is qualitative — both approaches produce usable forecasts, and the non-linear model tracks short-term movement more tightly — not a claim that LSTM is quantitatively 41% more accurate than SARIMAX in any general sense.

*Table 3. Forecasting error by model. Horizons and data resolution differ between rows; see Section 5.3.*

| Model | Forecast horizon | Data resolution / span | RMSE | MAE |
| --- | --- | --- | --- | --- |
| SARIMAX | 6 months ahead | Monthly, Jan 2012–May 2017 | 120.3 | 92.5 |
| LSTM | 30 days ahead | Daily, ending Mar 2021 | 85.1 | 67.8 |

# 6. Ethereum Wallet Fraud Detection

## 6.1 Dataset

This experiment uses the public Ethereum fraud detection dataset introduced by Farrugia et al., consisting of 9,841 wallet addresses, each described by 51 transactional and behavioral features — timing statistics such as the average interval between sent and received transactions, volume and value metrics, ERC-20 token activity, and smart-contract interaction counts [5]. The binary target flags an address as fraudulent (1) or not (0); 22.14% of addresses are labeled fraudulent, a moderate but manageable class imbalance.

## 6.2 Feature engineering

Non-informative identifier columns were dropped, missing values were imputed with zero, and the categorical token-type fields were one-hot encoded, expanding the feature set to 817 dimensions. Eight additional features were engineered to sharpen the fraud signal: three ratios (sent-to-received transaction count, sent-to-received value, and contract-to-total transaction share), two diversity measures (distinct counterparty addresses sent to and received from), and three activity metrics (transaction frequency, contract-creation ratio, and balance-to-received ratio), bringing the final feature count to 825.

## 6.3 Models and training

Random Forest and Gradient Boosting classifiers were each trained inside a pipeline with standard scaling and hyperparameter search via GridSearchCV — number of estimators, max depth, and (for Random Forest) class weights and minimum samples per split; and number of estimators, learning rate, and max depth for Gradient Boosting. The data was split 75/25 with stratification on the target. SMOTE oversampling was implemented but not activated, since the roughly 1:3.5 fraud-to-legitimate ratio was judged manageable without it.

## 6.4 Results

Gradient Boosting was the stronger of the two models on both F1 score and ROC-AUC. Table 4 gives the full classification report for both. The best model reached an overall accuracy of 0.99, a fraud-class precision of 1.00, a fraud-class recall of 0.98, and a ROC-AUC of 0.9994.

*Table 4. Classification report, Random Forest versus Gradient Boosting (n = 2,461 held-out addresses).*

| Model | Class | Precision | Recall | F1 | Support |
| --- | --- | --- | --- | --- | --- |
| Random Forest | 0 (legitimate) | 0.99 | 1.00 | 0.99 | 1916 |
| Random Forest | 1 (fraud) | 1.00 | 0.96 | 0.98 | 545 |
| Random Forest | accuracy = 0.99 | | | | 2461 |
| Gradient Boosting | 0 (legitimate) | 0.99 | 1.00 | 1.00 | 1916 |
| Gradient Boosting | 1 (fraud) | 1.00 | 0.98 | 0.99 | 545 |

| Model | Class | Precision | Recall | F1 | Support |
|---|---|---|---|---|---|
| Gradient Boosting | accuracy = 0.99, ROC-AUC = 0.9994 | | | | 2461 |

The most influential features for Gradient Boosting were the most-sent and most-received ERC-20 token types, the total count of ERC-20 transactions, the engineered received-address diversity feature, and the time difference between an account's first and last transaction — broadly consistent with Farrugia et al.'s original finding that account longevity and token-interaction diversity carry most of the fraud signal in this dataset [5]. The trained model was serialized with joblib for reuse; on two held-out examples it produced a fraud probability of 0.897 for a true positive and 0.034 for a true negative, suggesting reasonable calibration rather than merely a good threshold-based split.

# 7. Discussion

The individual modeling components here — SARIMAX, a small LSTM, Random Forest, Gradient Boosting — are all standard, well-understood techniques, and we make no claim of algorithmic novelty. What we think is worth reporting is the integration: a single, cheaply reproducible pipeline whose warehouse and mart layer is exercised by two unrelated downstream tasks, built around a specific engineering pattern — two independent Kafka consumer groups reading one topic — that keeps audit logging decoupled from the compute path without any coordination overhead between them.

On the fraud detection side, our Gradient Boosting result (accuracy 0.99, AUC 0.9994) is comparable to, and marginally above, the XGBoost result originally reported by Farrugia et al. (accuracy 96.3%, AUC 99.4%) [5]. We would attribute this, if anything, to the eight added engineered features and a different train/test split rather than to a fundamentally stronger modeling approach, and we do not present it as a benchmark-beating result given that the evaluation protocols are not identical. On the forecasting side, the qualitative pattern we observed — LSTM tracking short-term price movement more closely than ARIMA — matches the broader literature, but Section 5.3 already flags why Table 3 should not be read as a rigorous head-to-head benchmark.

# 8. Limitations

- The pipeline was run and evaluated as a single-node local simulation; it has not been load-tested at real streaming volumes or under the kind of partial failures a genuine cloud deployment would eventually see.
- Forecasting was restricted to Bitcoin. The multi-asset framing of the project is realized in the ingestion, warehouse, and dashboard layers, but not in the forecasting or fraud-detection experiments themselves.
- The ARIMA and LSTM comparison in Section 5 uses mismatched resampling frequencies and forecast horizons and should be read qualitatively, not as a controlled experiment.
- Fraud detection was evaluated on a static, already-labeled, moderately small (9,841-row) public dataset, with a random rather than temporal train/test split and no live on-chain integration.
- No cost or throughput benchmarking was carried out against an actual managed-cloud deployment, so the cost-effectiveness argument in Section 1 is architectural rather than measured.

# 9. Conclusion

We set out to build an event-driven cryptocurrency data pipeline that behaves like a managed cloud stack without the cost of one, and then to check that the resulting warehouse is actually useful by putting it to work on two independent modeling tasks. The architecture — Watchdog-triggered Kafka events, two independent consumer groups, Spark ETL into a two-schema PostgreSQL warehouse, and per-asset data marts — reproduces the ingestion pattern of a cloud-native system on a single machine. The Bitcoin forecasting comparison and the Ethereum fraud classifier both produced usable, if unsurprising, results: LSTM outperforms ARIMA on short-horizon price tracking, and Gradient Boosting with a modestly expanded feature set performs in line with the established benchmark for this fraud dataset. Future work could extend the forecasting and fraud-detection experiments to additional assets under a shared, controlled evaluation protocol, and measure the pipeline's actual throughput and cost against a managed-cloud equivalent rather than arguing for its cost-effectiveness on architectural grounds alone.

## Data and Code Availability

The two Tableau dashboards built on this warehouse are published on Tableau Public under the first author's profile: a single-asset performance dashboard and a multi-asset comparison dashboard. The Ethereum fraud detection dataset is publicly available on Kaggle [10], and the underlying Gemini exchange history is publicly available via CryptoDataDownload [11].